\documentclass[10pt,twocolumn,letterpaper]{article}
\usepackage[pagenumbers]{cvpr}
\usepackage{marvosym}
\usepackage{algorithm}
\usepackage{algpseudocode}
\usepackage{needspace}
\definecolor{cvprblue}{rgb}{0.21,0.49,0.74}
\usepackage[pagebackref,breaklinks,colorlinks,allcolors=cvprblue,hyperfootnotes=false]{hyperref}
\title{SpecWM: Turning Verification into Refinement for Streaming World Models}
\newcommand{\corrmark}{\href{mailto:mzyth@hust.edu.cn}{\textsuperscript{\Letter}}}
\author{%
\normalsize
\textbf{Jiaming Li}\textsuperscript{1,2,*}\hspace{7pt}
\textbf{Lingzhen Li}\textsuperscript{1,*}\hspace{7pt}
\textbf{Xiaze Yang}\textsuperscript{1}\hspace{7pt}
\textbf{Chenyu Zhu}\textsuperscript{1}\hspace{7pt}
\textbf{Kaiqi Liu}\textsuperscript{3}\hspace{7pt}
\textbf{Xiaotong Zhao}\textsuperscript{2}\\[0pt]
\normalsize
\textbf{Songlian Li}\textsuperscript{2}\hspace{7pt}
\textbf{Toyota Li}\textsuperscript{2}\hspace{7pt}
\textbf{Alan Zhao}\textsuperscript{2}\hspace{7pt}
\textbf{Haobin Zhong}\textsuperscript{2}\hspace{7pt}
\textbf{Yue Ma}\textsuperscript{4}\hspace{7pt}
\textbf{Zhiyuan Ma}\textsuperscript{1}\corrmark\\[4pt]
\small
\textsuperscript{1}Huazhong University of Science and Technology\hspace{7pt}
\textsuperscript{2}Tencent\\[0pt]
\small
\textsuperscript{3}Peking University\hspace{7pt}
\textsuperscript{4}The Hong Kong University of Science and Technology\\[1pt]
\footnotesize
\textsuperscript{*}Equal contribution.\quad\corrmark\,Corresponding author.
}
\hypersetup{
  pdftitle={SpecWM: Turning Verification into Refinement for Streaming World Models},
  pdfauthor={Jiaming Li, Lingzhen Li, Xiaze Yang, Chenyu Zhu, Kaiqi Liu, Xiaotong Zhao, Songlian Li, Toyota Li, Alan Zhao, Haobin Zhong, Yue Ma, Zhiyuan Ma}
}

\makeatletter
\def\ps@specwmpreprint{%
  \let\@oddhead\@empty
  \let\@evenhead\@empty
  \def\@oddfoot{\normalfont\hbox to\textwidth{\rlap{\small Preprint}\hfil\thepage\hfil}}%
  \let\@evenfoot\@oddfoot
}
\patchcmd{\@maketitle}{\vskip .375in}{\vskip 8pt}{}{}
\patchcmd{\@maketitle}{\vspace*{24pt}}{\vspace*{8pt}}{}{}
\patchcmd{\@maketitle}{\vspace*{12pt}}{\vspace*{2pt}}{}{}
\g@addto@macro\@maketitle{\vspace{-10pt}
\begin{minipage}{\textwidth}
\centering
\captionsetup{hypcap=false}
\includegraphics[width=0.98\textwidth]{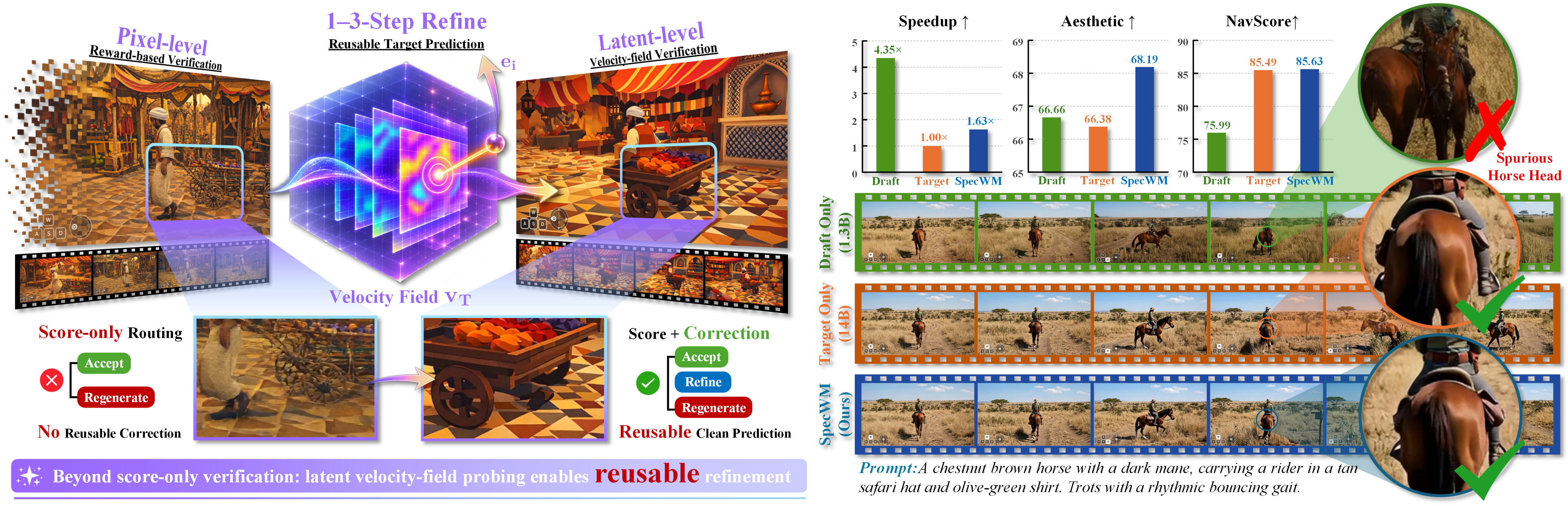}
\captionof{figure}{\textbf{Turning verification into refinement.} Left: latent verification supplies both a score and a reusable correction. Right: SpecWM improves aesthetics and retains target-level navigation control at 1.63$\times$ speedup over Target-only on WBench Navigation.}
\label{fig:motivation}
\end{minipage}
\par\vspace{4pt}
}
\makeatother
\newcommand{\mainparagraph}[1]{\par\Needspace{2\baselineskip}\noindent\textbf{#1}\enspace\ignorespaces}
\begin{document}
\raggedbottom
\maketitle
\thispagestyle{specwmpreprint}
\begingroup
\renewcommand{\endabstract}{\par}
\begin{abstract}
Recently, video world models have advanced interactive generation, with large models offering higher quality and smaller models enabling faster inference. Existing speculative methods combine their strengths through fast draft generation and selective target regeneration guided by rewards on decoded frames. However, verification covers only the external scorer's evaluation dimensions and adds decoding and scoring costs. Moreover, these scores provide no correction, requiring costly target regeneration for rejected drafts. To address these limitations, we propose \textbf{Speculative World Model (SpecWM)}, a unified, training-free inference framework for streaming world models. It verifies draft chunks using the target's latent velocity field. Verification also yields a reusable correction; two thresholds select Accept, Refine, or Reject. Specifically, one-step refinement reuses it to improve perceptual quality, while three-step refinement strengthens camera following and trajectory consistency. These findings motivate transition-aware refinement. On WBench Navigation, SpecWM improves visual quality at 1.63$\times$ speedup over the target and raises NavScore from the draft's 75.99 to 85.63, achieving the best quality-speed-control trade-off. Additionally, our LongSpecBench shows sustained quality and control over long, multi-turn rollouts, while experiments on Hy-Nomas validate SpecWM across backbones and denoising budgets.
\end{abstract}

\endgroup
\makeatletter
\if@firstcolumn\newpage\fi
\makeatother
\section{Introduction}
\label{sec:intro}
Streaming world models enable users to explore generated environments that evolve in response to their actions~\cite{lingbot,solarwm}. Such interaction requires high visual quality, accurate control following, and rapid responses to changing inputs. However, larger models offer stronger quality and control at substantial computational cost, whereas smaller models generate faster at the expense of visual quality and accuracy in following the controls (Fig.~\ref{fig:motivation}, right).

Existing reward-based speculative methods combine these strengths through fast draft generation and selective target regeneration. Rewards computed on decoded draft frames determine which chunks to accept and which to regenerate with the target. Setting up this verifier requires choosing a reward model, aggregating frame scores, and calibrating an acceptance threshold~\cite{sdvg}. However, verification is limited to the attributes captured by the external scorer~\cite{imagereward}. Decoding and scoring also add runtime costs per chunk. Moreover, these scores provide no direct refinement signal to improve the draft itself (Fig.~\ref{fig:motivation}, left). Consequently, every rejected draft requires computationally expensive regeneration by the target through its full sampling process, reducing the savings from fast draft generation. \emph{How can verification itself provide a reusable correction to improve the trade-off among speed, visual quality, and control in speculative generation for streaming world models?}

We propose \textbf{Speculative World Model (SpecWM)}, a unified, training-free inference framework that turns verification into refinement for streaming autoregressive (AR) world models. SpecWM replaces decoded-frame reward scoring with a single final-stage evaluation of the target's latent velocity field $v$. This evaluation yields a clean prediction under the same history and controls. Its difference from the draft provides a verification score. Moreover, the same prediction supplies a reusable correction. Based on this score, SpecWM uses two thresholds to select Accept, Refine, or Reject to allocate target computation across chunks.

One-step refinement directly reuses the prediction from verification, without further target denoising. Three-step refinement starts at a higher noise level and runs three consecutive target denoising steps. With verification fixed, our uniform-refinement experiments reveal complementary effects (Table~\ref{tab:main_speculative}). One-step refinement improves imaging quality and perceptual preference over the draft, with limited improvement in navigation control. Three-step refinement strengthens camera following and trajectory consistency, but lowers perceptual preference and aesthetics relative to one step; imaging quality remains similar. Image diffusion studies link early denoising to coarse structure and later steps to fine detail~\cite{yi2024mechanism}. Our results suggest a similar distinction: final-stage correction improves appearance, while earlier intervention allows greater trajectory revision.

These findings motivate \textbf{transition-aware refinement}. For chunks routed to Refine, SpecWM uses three steps at a control change, two in the next chunk, and one thereafter. This schedule improves control near transitions while favoring perceptual quality and speed during continued motion. Throughout this schedule, verification remains fixed at one final-stage probe per draft chunk. Additionally, the draft generates the next chunk while the target verifies the current one, thereby reducing latency.

Our main contributions are summarized as follows:
\begin{itemize}[leftmargin=*,nosep]
\item \textbf{SpecWM.} We propose a unified, training-free framework that turns verification into refinement. With transition-aware refinement, SpecWM achieves a 1.63$\times$ generation-loop speedup over Target-only on WBench Navigation~\cite{wbench}, improving imaging quality and perceptual preference while maintaining target-level navigation control.
\item \textbf{LongSpecBench.} We construct a benchmark of long, multi-turn rollouts with revisits to jointly evaluate generation quality, speed, and sustained controllability. SpecWM leads the evaluated methods in world-model quality (WMQ) and quality-efficiency score (QES).
\item \textbf{Beyond model scale.} SpecWM extends from model-scale to denoising-budget differences on Hy-Nomas~\cite{hynomas}, where draft and target share the same weights. It outperforms Draft-only and SDVG-style across four human-evaluated dimensions, achieving near-target preference at a 1.33$\times$ generation-loop speedup over Target-only.
\end{itemize}

\begin{figure*}[t]
\centering
\includegraphics[width=\textwidth]{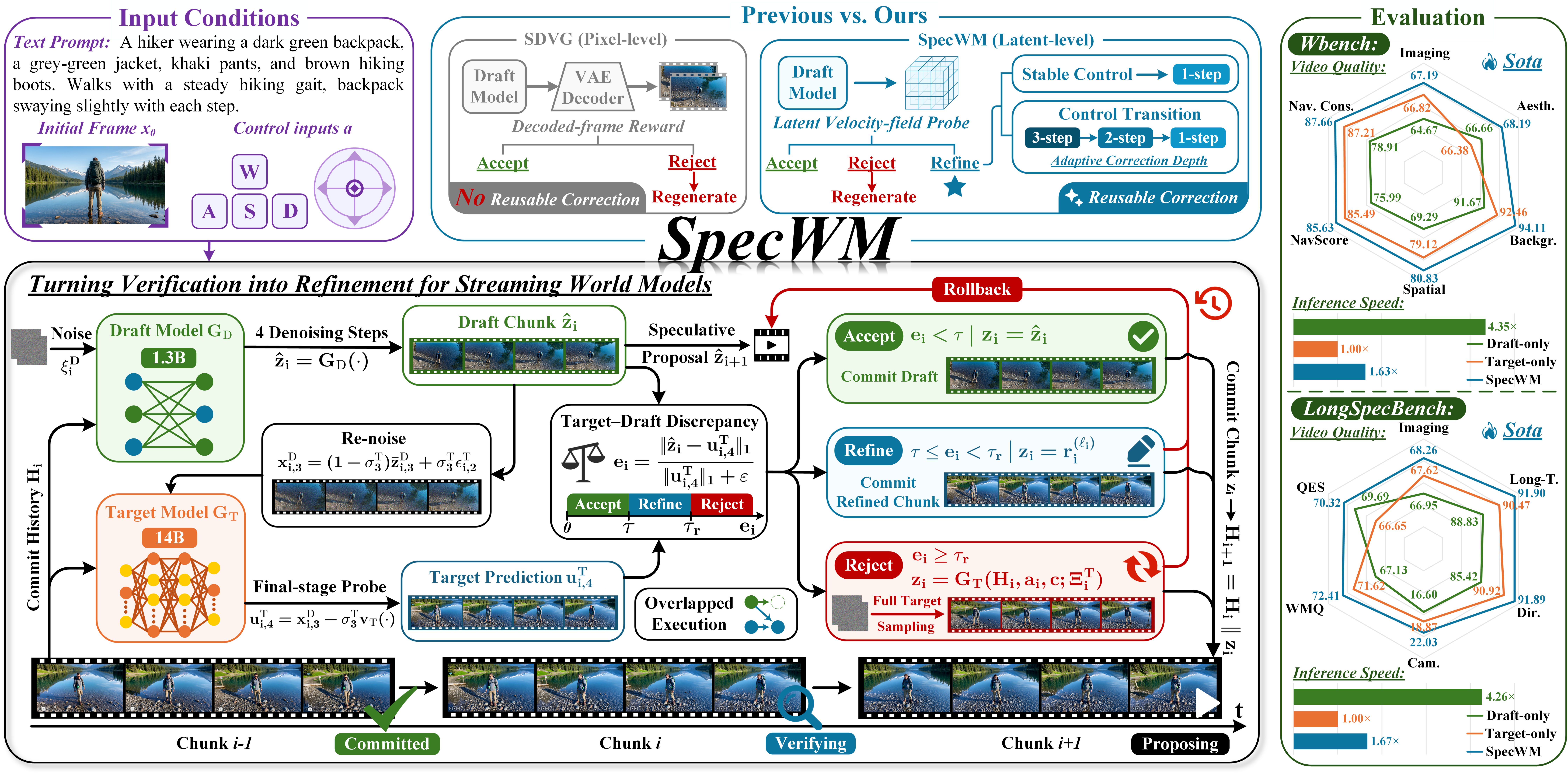}
\caption{\textbf{SpecWM overview.} Top: decoded-frame reward verification and SpecWM's latent verification with transition-aware refinement. Main: a fixed final-stage target probe supplies a score and a reusable clean prediction. Two thresholds select Accept, Refine, or Reject. One-step Refine reuses the prediction; longer corrections follow the $3\to2\to1$ schedule near control transitions. Fig.~\ref{fig:overlap} details overlapped execution and re-proposal. Right: selected quality and control scores, and speedups of Draft-only, Target-only, and SpecWM on WBench Navigation and LongSpecBench (Tables~\ref{tab:main_speculative} and~\ref{tab:longspec}), allowing direct comparison across both evaluation settings.}
\label{fig:pipeline}
\end{figure*}

\Needspace{5\baselineskip}
\section{Related Work}
\label{sec:related}

\mainparagraph{Autoregressive video generation and world models.}
Diffusion models~\cite{ddpm}, latent representations~\cite{ldm}, and transformers~\cite{dit} underpin video synthesis~\cite{videodiffusion,svd,videocrafter2,cogvideox,animatediff}. Building on this foundation, camera and motion conditioning enable controllable generation~\cite{cameractrl,motionctrl}, while flexible temporal conditioning incorporates generation history~\cite{diffusionforcing,historyguidance}. For streaming generation, causal distillation reduces latency~\cite{causvid,cfpp}, and training on self-generated histories addresses the training-inference mismatch~\cite{selfforcing}. Longer rollouts further require managing accumulated errors through long-video tuning, cache refreshes, and rolling denoising~\cite{longlive,rollingforcing}. These developments support interactive world models that respond to user actions~\cite{genie,lingbot,solarwm,hynomas}. Evaluation likewise extends from video quality in VBench~\cite{vbench} to multi-turn interaction in WBench~\cite{wbench}. SpecWM addresses computation allocation during streaming inference to balance visual quality, speed, and control.

\mainparagraph{Caching and efficient inference.}
Efficient samplers~\cite{ddim,dpmsolver}, distillation~\cite{progressivedistill,dmd}, and consistency models~\cite{consistency} reduce denoising steps. Beyond reducing steps, caching exploits redundancy in features, attention, and guidance~\cite{deepcache,fastercache}. Adaptive reuse responds to estimated output changes or chunk-level variation~\cite{teacache,flowcache}. For world models, selective token recomputation~\cite{worldcache} and history-guided caching with residual correction~\cite{arcache} address reuse across evolving content. System optimizations combine caching with sparse attention, quantization, and kernel fusion~\cite{solengine}. These approaches lower the cost of running a generator by reusing internal computations or reducing sampling steps. SpecWM instead allocates computation according to the draft chunk's discrepancy from the target. Accept, Refine, and Reject provide different levels of target intervention. This allocation applies to pairings with different model scales or denoising budgets, including shared-weight configurations.

\mainparagraph{Speculative verification and refinement.}
Speculative decoding combines draft proposals with target verification~\cite{leviathan,chenspec}; feature prediction and multiple decoding heads improve proposal generation~\cite{eagle,medusa}. DSpark uses confidence and serving load to schedule token verification~\cite{dspark}. In video generation, SDVG uses ImageReward~\cite{imagereward} on decoded draft frames to select acceptance or target regeneration~\cite{sdvg}. However, this verifier incurs decoding and scoring costs, and its coverage depends on the external scorer. Moreover, the scores guide selection without supplying a direct correction. Separately, other methods allocate computation within one generator: Speculative Forcing freezes converged tokens~\cite{specforcing}, while Surprise Forcing skips steps according to chunk difficulty~\cite{surpriseforcing}. SpecWM instead verifies draft chunks using the target's latent velocity prediction. The resulting clean estimate can correct the draft, adding refinement between acceptance and full regeneration.

\Needspace{5\baselineskip}
\section{Preliminaries}
\label{sec:prelim}

\mainparagraph{Controlled autoregressive generation.}
Let $z_i$ denote chunk $i$'s clean latent, $H_i=(z_0,\ldots,z_{i-1})$ the committed history, $a_i$ the camera controls, and $c$ the initial-image and text conditions. For $N$ chunks, the generator samples
\begin{equation}
z_i\sim G(\cdot\mid H_i,a_i,c),\qquad i=0,\ldots,N-1.
\label{eq:ar}
\end{equation}
The selected output is appended to the history. For speculative generation, we pair frozen generators $G_D$ and $G_T$ with different model scales or denoising budgets. They share a latent representation, chunk length, and control interface, and condition on the same history and inputs.

\mainparagraph{Velocity-based denoising.}
Following flow matching and rectified flow~\cite{flowmatching,rectifiedflow}, we use a linear interpolation between a clean latent $z$ and noise $\epsilon$. At noise level $\sigma$, the noisy latent is $x_\sigma=(1-\sigma)z+\sigma\epsilon$. The model predicts a latent velocity field $v$, from which we obtain the clean estimate
\begin{equation}
\widetilde z=x_\sigma-\sigma v(x_\sigma,\sigma;H_i,a_i,c).
\label{eq:clean}
\end{equation}
For the LingBot implementation~\cite{lingbot}, we use four native denoising stages $k\in\{0,1,2,3\}$, with decreasing noise levels $1\geq\sigma_0>\sigma_1>\sigma_2>\sigma_3>0$. At intermediate stages $k<3$, the clean estimate is re-noised to $\sigma_{k+1}$ for the next update. At the final stage, it is returned.

\section{SpecWM}
\label{sec:method}
SpecWM turns target verification into reusable refinement (Fig.~\ref{fig:pipeline}, main panel). A final-stage velocity prediction yields a clean candidate, whose discrepancy from the draft provides a verification score. Two thresholds then select Accept, Refine, or Reject to allocate target computation. For chunks routed to Refine, control transitions determine the correction entry point. With separate models, draft generation overlaps target verification to reduce latency. Both roles then use the selected output as shared context.

\subsection{Verification and Reusable Refinement}
\label{sec:verify}
We describe SpecWM using LingBot's four-stage sampler. Draft predictions are aligned with target noise levels. For shared-weight pairing, we instead subdivide the draft's final update along the target grid (Sec.~\ref{sec:exp_hynomas}).

\mainparagraph{Draft proposal and alignment.}
The draft generates $\widehat z_i=G_D(H_i,a_i,c;\xi_i^D)$, where $\xi_i^D$ denotes its sampling randomness, and retains the clean prediction from each denoising stage. Draft and target can have different noise schedules. We therefore interpolate the draft's clean predictions by noise level to obtain $\bar z^D_{i,k}$ at target level $\sigma_k^T$. To form target inputs, we re-noise these estimates using $\Xi_i^T=(\eta_i^T,\epsilon^T_{i,0:2})$, which contains the target's initial noise and three re-noising draws. The resulting aligned states are given by
\begin{align}
x^D_{i,0}&=\eta_i^T,\qquad x^D_{i,4}=\widehat z_i,\nonumber\\
x^D_{i,k}&=(1-\sigma_k^T)\bar z^D_{i,k}
              +\sigma_k^T\epsilon^T_{i,k-1},
\quad k=1,2,3.
\label{eq:mapping}
\end{align}
These states provide the input for verification and the starting points for longer corrections. Interpolation and random-state details are given in the supplementary material.

\mainparagraph{One probe, two outputs.}
Using the aligned final-stage state, we evaluate the target at $k=3$ to obtain
\begin{equation}
u^T_{i,4}=x^D_{i,3}-\sigma_3^T v_T(x^D_{i,3},\sigma_3^T;H_i,a_i,c).
\label{eq:target_clean}
\end{equation}
We then compare this clean prediction with the draft output:
\begin{equation}
e_i=\frac{\|\widehat z_i-u^T_{i,4}\|_1}{\|u^T_{i,4}\|_1+\varepsilon},
\qquad \varepsilon=10^{-6}.
\label{eq:score}
\end{equation}
The norms cover valid latent elements. The score $e_i$ guides routing, while the prediction $u^T_{i,4}$ can be committed as a one-step correction without further target denoising. Thus, verification directly provides a refinement candidate. Keeping this probe fixed gives all refinement lengths the same scoring rule. Its low cost and observed score variation across chunks motivate the final-stage choice (Sec.~\ref{sec:exp_probe}).

\subsection{Threshold-Based Computation Allocation}
\label{sec:routing}
Given the discrepancy score $e_i$, two thresholds $\tau<\tau_r$ determine how the target intervenes. Let $r_i^{(\ell_i)}$ denote a corrected chunk obtained with $\ell_i$ target denoising steps, as defined in Sec.~\ref{sec:depth}. The selected output is
\begin{equation}
z_i=
\begin{cases}
\widehat z_i,&e_i<\tau\quad\text{(Accept)},\\
r_i^{(\ell_i)},&\tau\leq e_i<\tau_r\quad\text{(Refine)},\\
G_T(H_i,a_i,c;\Xi_i^T),&e_i\geq\tau_r\quad\text{(Reject)}.
\end{cases}
\label{eq:route}
\end{equation}
Accept retains the draft when the discrepancy is small. Refine corrects intermediate discrepancies, reusing the probe candidate for the shortest correction or running a longer target chain. Reject runs the target's full sampler to regenerate the chunk. Thus, the thresholds choose the action, while the transition schedule selects where refinement begins.

\mainparagraph{Burst fallback.}
\label{sec:backoff}
Repeated rejections incur draft and probe costs before each full regeneration. To avoid these recurring costs, SpecWM switches to target-only generation after $m=2$ consecutive Reject decisions. It generates the next $n=4$ chunks directly with the target, up to the end of the rollout. Fallback skips draft proposals and verification probes; target outputs are committed directly. Both models append these outputs to their histories before normal routing resumes with the rejection counter reset. We denote these target-only chunks by T in the routing statistics, separately from chunks regenerated after a Reject decision.

\subsection{Transition-Aware Refinement}
\label{sec:depth}
Once a chunk is routed to Refine, the remaining choice is its correction length. Our uniform-refinement study reveals complementary effects (Table~\ref{tab:main_speculative}). One-step correction improves perceptual quality over the draft. Three-step correction improves camera following and trajectory consistency, but lowers perceptual preference and aesthetics relative to one step. We therefore allocate longer corrections near control transitions and shorter corrections during continued motion, with verification fixed.

\mainparagraph{Refinement entry point.}
Control transitions determine how far back along the denoising trajectory correction begins. For LingBot, this gives $\ell_i=3$ steps at a transition, two in the next chunk, and one thereafter. The initial chunk is treated as a transition. The schedule follows chunk positions, regardless of earlier routing decisions. Controls or navigation-turn metadata identify transitions, as detailed in the supplementary material.

\mainparagraph{Executing the correction.}
For $\ell_i=1$, we directly reuse $r_i^{(1)}=u^T_{i,4}$. For longer corrections, $\ell_i\in\{2,3\}$, we start from the aligned state $q_{4-\ell_i}=x^D_{i,4-\ell_i}$ and run consecutive target updates for $k=4-\ell_i,\ldots,3$:
\begin{align}
\widetilde z_k&=q_k-\sigma_k^T v_T(q_k,\sigma_k^T;H_i,a_i,c),\nonumber\\
q_{k+1}&=
\begin{cases}
(1-\sigma_{k+1}^T)\widetilde z_k+\sigma_{k+1}^T\epsilon^T_{i,k},&k<3,\\
\widetilde z_k,&k=3.
\end{cases}
\label{eq:native_update}
\end{align}
The final state $q_4$ becomes $r_i^{(\ell_i)}$. Each update uses the preceding target output under the fixed history. The initial probe selects the route, and the resulting correction is committed once; the score is not recomputed during refinement. Fixed-length variants replace the schedule with a constant $\ell_i$ across all chunks routed to Refine.

A longer correction starts at a higher noise level and lets the target revise more of the denoising trajectory, paralleling the role of entry noise in diffusion editing~\cite{sdedit}. Across backbones, the policy therefore selects a correction interval on the native noise grid, which in turn determines the number of target updates (Sec.~\ref{sec:exp_hynomas}).

\subsection{Overlapped Speculative Execution}
\label{sec:pipeline}
\begin{figure}[!t]
\centering
\includegraphics[width=\columnwidth]{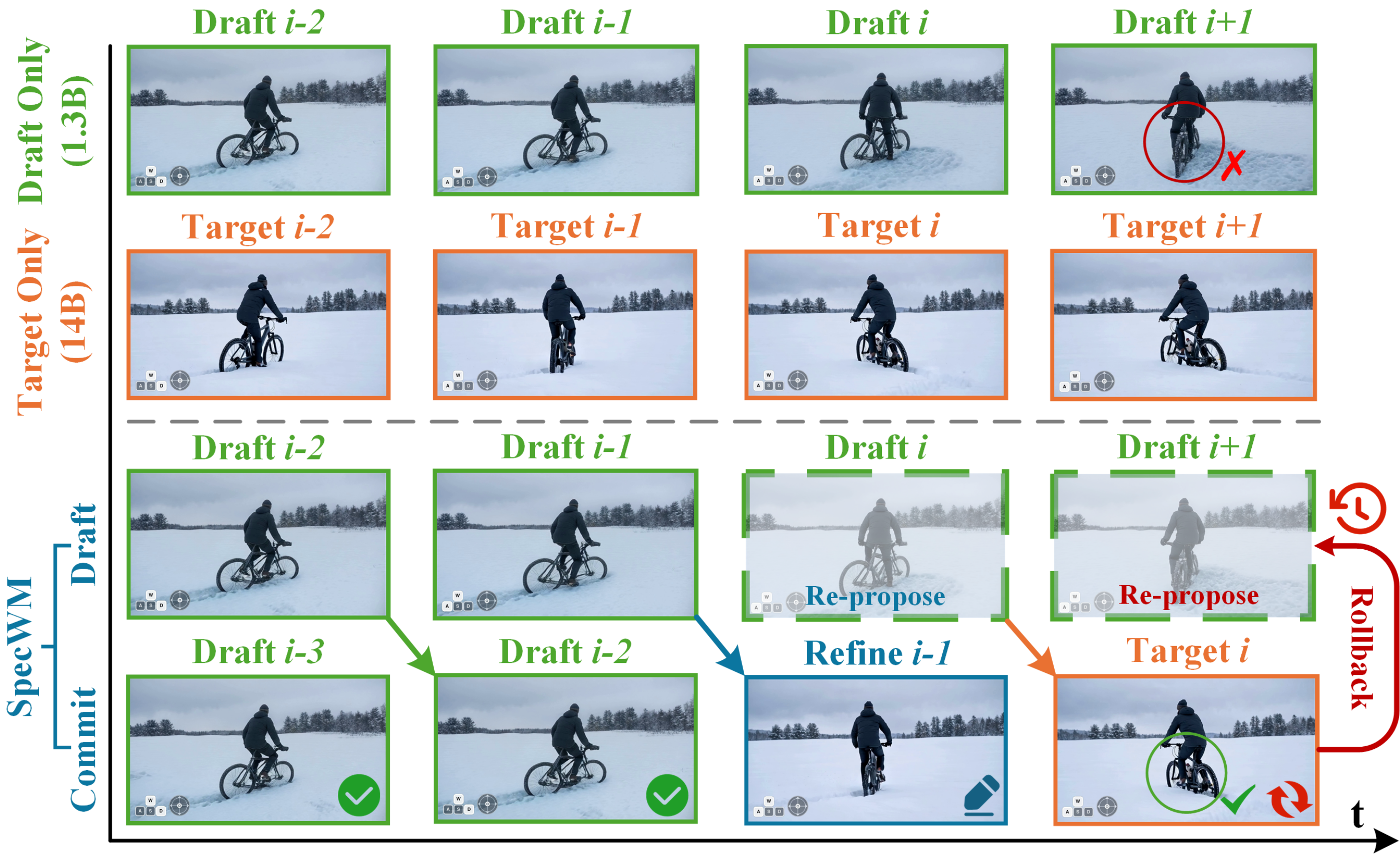}
\caption{\textbf{Overlapped speculative execution.} Top: Draft-only and Target-only rollouts. Bottom: SpecWM drafts the next chunk while verifying the current one. Accept preserves the next proposal; committing a refined or regenerated chunk invalidates it. Dashed proposals are discarded and re-proposed after rollback, using the updated history that includes the committed output.}
\label{fig:overlap}
\end{figure}

During target verification of chunk $i$, the draft generates chunk $i+1$ under the tentative history $H_i\Vert\widehat z_i$ (Fig.~\ref{fig:overlap}). This LingBot pipeline therefore reduces latency while still verifying one candidate at a time ($K=1$).

\mainparagraph{Retain or re-propose.}
Accept preserves this history, so the next proposal can be retained. Refine or Reject replaces $\widehat z_i$ with $z_i$, making the next proposal's history outdated. The draft then restores its state from before proposing chunk $i$, appends $z_i$, and re-proposes chunk $i+1$. The target also appends $z_i$. Fallback chunks update both histories.

To maintain consistent histories, both LingBot models commit the selected latent through clean-context forward passes into separate key-value (KV) caches. Temporary state is restored after probing or correction, before the selected output is committed. With shared weights, Hy-Nomas instead executes both roles sequentially using one cache with native history noise, updated only after output selection. The supplementary material details both implementations and their cache updates.

\mainparagraph{Denoising cost and runtime.}
\label{sec:cost}
For LingBot, Accept and one-step Refine each use one target probe. Two- and three-step Refine add $\ell_i$ target denoising calls after that probe. Reject runs the complete four-step target sampler; fallback runs it without a probe. These counts describe target denoising work; draft generation and cache updates add separate costs. Because overlap depends on retaining the next proposal, we measure the full generation loop, including cache updates, restoration, re-proposal, and communication. We therefore report speedup alongside A/F/R/T shares to relate measured runtime to routing decisions.

\section{Experiments}
\label{sec:experiments}
We evaluate SpecWM on WBench Navigation and ablate refinement, routing, and execution. LongSpecBench then tests sustained control and scene consistency over long rollouts, while Hy-Nomas examines shared-weight pairing through different denoising budgets.

\subsection{Experimental Setup}
\label{sec:exp_setup}

\begin{table*}[!t]
\centering
\caption{Results on the full 158-case WBench Navigation split with the LingBot-World~2.0 1.3B draft and 14B target. Quality and control scores are multiplied by 100. Parentheses specify refinement length; $3\to2\to1$ is the full transition-aware schedule. Speedup is relative to Target-only. A/F/R/T gives chunk percentages for Accept, Refine, Reject, and target-only fallback. All SpecWM variants retain fallback and overlapped execution. \textbf{Bold} and \underline{underline} mark the best and second-best values in each performance column.}
\label{tab:main_speculative}
\scriptsize
\setlength{\tabcolsep}{2.6pt}
\renewcommand{\arraystretch}{1.16}
\resizebox{\textwidth}{!}{
\begin{tabular}{lcc@{\hspace{6pt}}cccc@{\hspace{6pt}}cc@{\hspace{6pt}}ccc}
\toprule
& Efficiency & Routing & \multicolumn{4}{c}{Visual $\uparrow$} & \multicolumn{2}{c}{Consistency $\uparrow$} & \multicolumn{3}{c}{Camera $\uparrow$} \\
\cmidrule(lr){2-2}\cmidrule(lr){3-3}\cmidrule(lr){4-7}\cmidrule(lr){8-9}\cmidrule(lr){10-12}
Method & Speedup $\uparrow$ & A/F/R/T (\%) & HPSv3 & Imaging & Aesth. & Plaus. & Backgr. & Spatial & NavScore & Nav. Acc. & Nav. Cons. \\
\midrule
\multicolumn{12}{l}{\textit{Single-model references}} \\
Target-only (14B) & 1.00$\times$ & \textemdash & 74.57 & 66.82 & 66.38 & \textbf{61.69} & 92.46 & 79.12 & \underline{85.49} & \textbf{83.76} & \underline{87.21} \\
Draft-only (1.3B) & \textbf{4.35}$\times$ & \textemdash & 72.49 & 64.67 & 66.66 & 60.07 & 91.67 & 69.29 & 75.99 & 73.07 & 78.91 \\
\midrule
\multicolumn{12}{l}{\textit{Uniform policies}} \\
Accept-all & \underline{2.50}$\times$ & 100/0/0/0 & 72.71 & 64.67 & 66.16 & 60.07 & 91.67 & 69.29 & 75.94 & 73.13 & 78.74 \\
Refine-all (1-step) & 1.78$\times$ & 0/100/0/0 & \textbf{77.80} & 66.35 & \underline{68.10} & 61.44 & \underline{94.04} & 70.65 & 75.92 & 72.69 & 79.15 \\
Refine-all (2-step) & 1.26$\times$ & 0/100/0/0 & 76.31 & \underline{66.95} & 67.48 & 61.37 & 93.49 & 69.97 & 77.42 & 74.07 & 80.76 \\
Refine-all (3-step) & 1.10$\times$ & 0/100/0/0 & 74.26 & 66.38 & 65.83 & 60.93 & 92.64 & \underline{80.69} & 85.17 & 83.17 & 87.16 \\
\midrule
\multicolumn{12}{l}{\textit{Selective generation}} \\
SDVG-style & 1.42$\times$ & 68/0/32/0 & 73.98 & 66.04 & 66.51 & 60.58 & 92.58 & 67.25 & 75.92 & 72.17 & 79.67 \\
SpecWM (1-step) & 1.84$\times$ & 61/21/10/8 & 76.73 & 65.91 & 67.19 & 60.62 & 93.60 & 70.09 & 76.39 & 72.88 & 79.90 \\
SpecWM (2-step) & 1.21$\times$ & 57/19/12/12 & 75.39 & 65.45 & 66.81 & 60.27 & 93.24 & 72.78 & 77.33 & 74.03 & 80.62 \\
SpecWM (3-step) & 1.13$\times$ & 58/20/11/11 & 73.84 & 65.41 & 66.42 & 60.25 & 92.16 & 78.75 & 84.63 & 82.58 & 86.68 \\
\addlinespace[2pt]
\textbf{SpecWM ($3\to2\to1$)} & 1.63$\times$ & 68/13/10/9 & \underline{77.66} & \textbf{67.19} & \textbf{68.19} & \underline{61.51} & \textbf{94.11} & \textbf{80.83} & \textbf{85.63} & \underline{83.59} & \textbf{87.66} \\
\bottomrule
\end{tabular}
}
\end{table*}

\mainparagraph{Models and generation.}
We use the distilled LingBot-World~2.0 causal-fast models~\cite{lingbot} for WBench and LongSpecBench: a 1.3B draft and a 14B target with native four-stage sampling. WBench videos contain 253 frames at 16\,fps (about 16 seconds), generated in 16 chunks of four latent frames. Resolution follows the input aspect ratio with a maximum pixel area of 480$\times$832. We use a context window of 18 latent frames and six sink frames. Methods share inputs, seeds, and GPU type. Sequence parallelism uses four GPUs per model: four for single-model baselines and eight for draft-target overlap.

\mainparagraph{Benchmarks.}
Main comparisons and ablations use the full 158-case WBench Navigation split~\cite{wbench} with shared scenes and controls. Each metric uses its applicable cases and valid outputs. LongSpecBench and Hy-Nomas-Bench are detailed in Secs.~\ref{sec:exp_longspec} and~\ref{sec:exp_hynomas}.

\mainparagraph{Metrics.}
Table~\ref{tab:main_speculative} groups metrics into visual quality, consistency, and camera control. HPSv3~\cite{hpsv3} follows WBench normalization; Spatial measures consistency upon returning to the initial viewpoint~\cite{wbench}. NavScore averages navigation accuracy and trajectory consistency, each computed over valid cases using estimated camera poses. Quality and control scores are multiplied by 100; Speedup and probe discrepancies retain their original scales.

\mainparagraph{Generation speedup.}
Speedup is the reciprocal of the mean per-case runtime ratio to Target-only. Across all three benchmarks, timing covers the full generation loop, including state updates, communication, and any rollback, decoding, or scoring required by routing. Model loading and final output decoding are excluded.

\mainparagraph{Compared methods.}
Target-only and Draft-only provide quality and speed references. Accept-all probes and retains every draft, while Refine-all corrects every chunk. SDVG-style adapts decoded-frame reward routing~\cite{sdvg} to the same model pair. LingBot-based SpecWM uses fixed refinement lengths (1/2/3-step) or the $3\to2\to1$ schedule. These variants share the score, thresholds, fallback, and overlap to isolate refinement length.

\mainparagraph{Policy settings.}
All latent-probe variants on LingBot use final-stage verification. For WBench, Accept-all discrepancies on 20 calibration cases set $\tau=0.2381$ and $\tau_r=0.2704$ at the 70th and 90th percentiles, fixed throughout evaluation. The full policy uses $3\to2\to1$ refinement, burst fallback $(m,n)=(2,4)$, and overlap with $K=1$.

\subsection{Main Results on WBench}
\label{sec:exp_main}

\begin{figure}[!b]
\centering
\vspace{-22pt}
\includegraphics[width=\columnwidth]{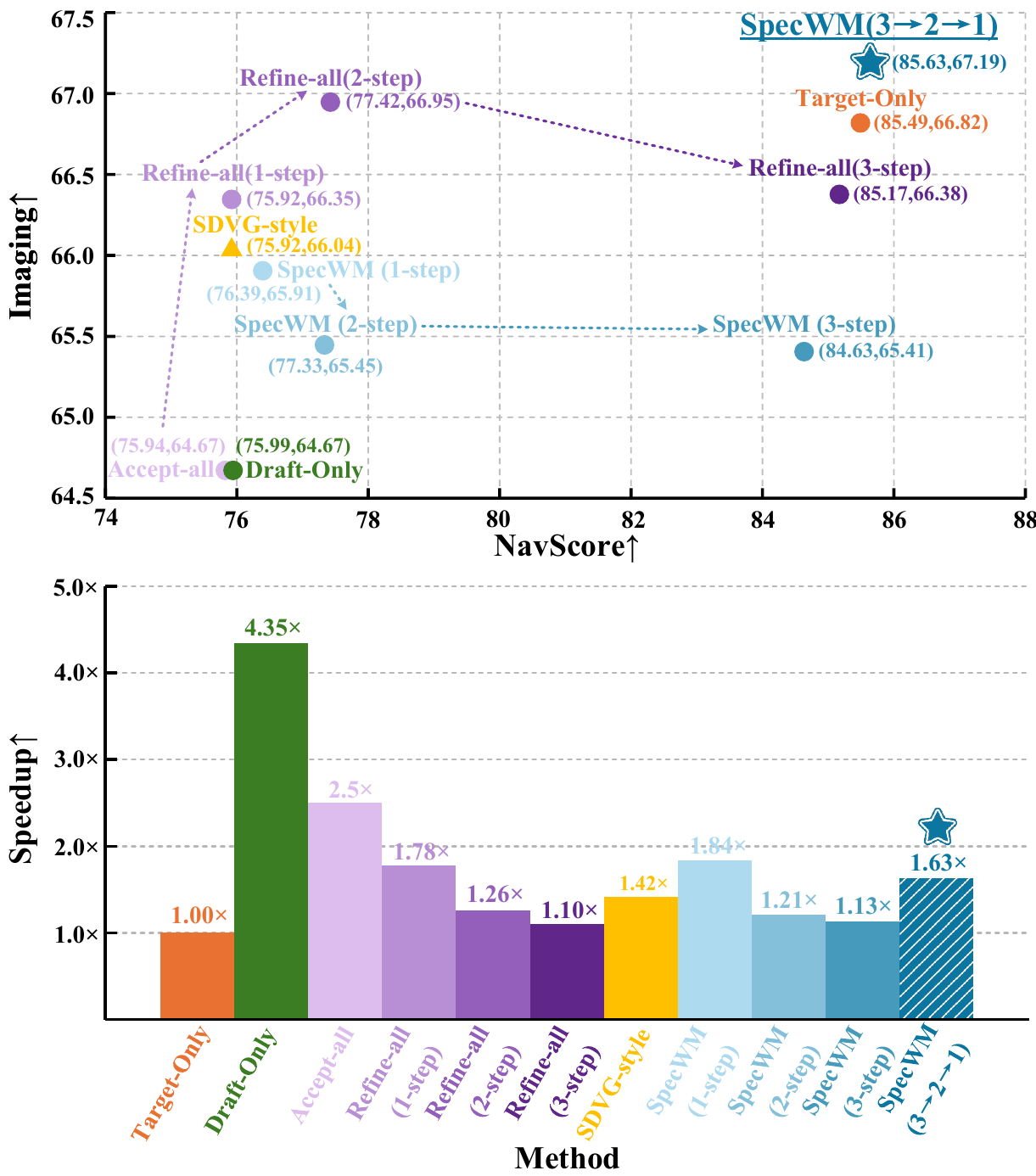}
\caption{\textbf{WBench quality-speed-control comparison.} Top: Imaging versus NavScore; dotted arrows indicate increasing refinement length within each policy family. Bottom: generation-loop speedup over Target-only. Longer fixed refinements improve navigation but reduce speed. In comparison, SpecWM achieves the highest Imaging and NavScore at 1.63$\times$ speedup (Table~\ref{tab:main_speculative}).}
\label{fig:pareto}
\end{figure}

\mainparagraph{Quality-speed-control trade-off.}
SpecWM achieves the best overall quality-speed-control trade-off among the evaluated methods (Table~\ref{tab:main_speculative}, Fig.~\ref{fig:pareto}). At 1.63$\times$ speedup over Target-only, it retains comparable NavScore (85.63 versus 85.49) and improves HPSv3 from 74.57 to 77.66. It also gains 9.64 NavScore points over Draft-only. Our selective target intervention thus recovers target-level control with lower latency and higher perceptual preference.

\mainparagraph{Effects of refinement length.}
Longer refinement favors control over perceptual preference. Increasing Refine-all from one to three steps raises NavScore from 75.92 to 85.17, but lowers HPSv3 from 77.80 to 74.26 and speedup from 1.78$\times$ to 1.10$\times$. Imaging quality remains similar. These complementary effects motivate allocating correction length around transitions. By selecting where and how long to refine, SpecWM therefore approaches both one-step perceptual preference and target-level control, while maintaining lower latency than uniform three-step correction.

\begin{figure*}[!t]
\centering
\includegraphics[width=\textwidth]{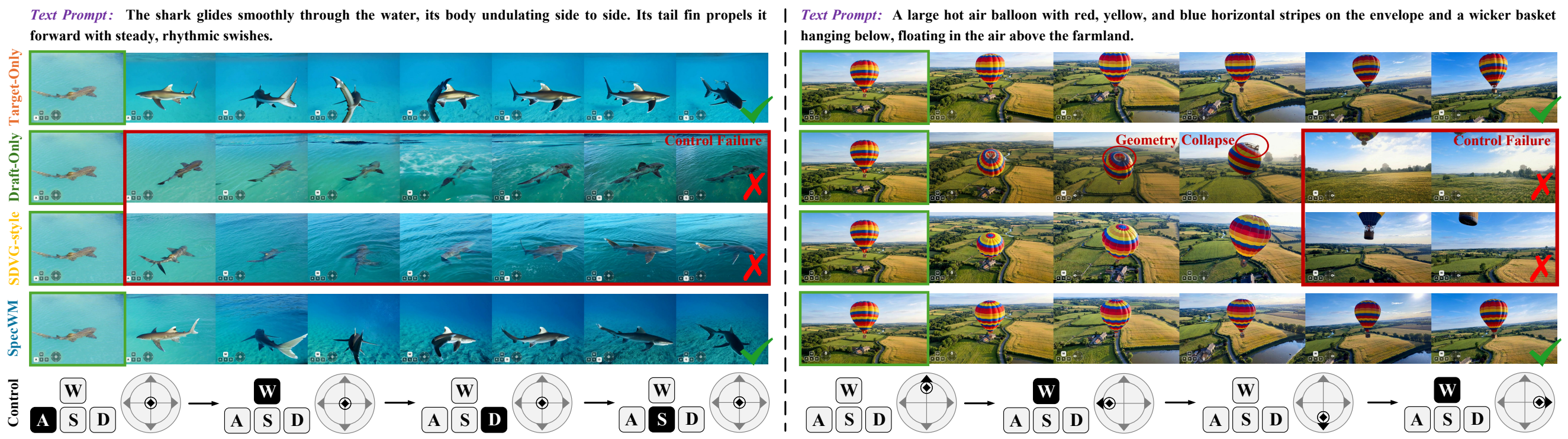}
\caption{\textbf{Qualitative comparison under changing controls.} Target-only, Draft-only, SDVG-style, and SpecWM share the inputs and control sequence in each case. Control inputs are shown below the frames. Highlighted regions mark control failures and geometry collapse. SpecWM follows the changing controls while preserving scene structure across both sequences.}
\label{fig:qualitative}
\end{figure*}

\mainparagraph{Refinement reduces regeneration.}
Equal acceptance rates need not yield equal efficiency. Both SDVG-style and SpecWM accept 68\% of chunks, but SpecWM refines 13\%, leaving 19\% for full target generation, including fallback, versus 32\% for SDVG-style. Refinement therefore reduces the share requiring a full target run. Together with latent verification and overlapped execution, SpecWM raises speedup from 1.42$\times$ to 1.63$\times$ while improving visual quality and control. Fig.~\ref{fig:qualitative} illustrates its preservation of scene structure and response to changing controls in both cases.

\subsection{Ablation Studies and Analysis}
\label{sec:exp_route}
Table~\ref{tab:routing_ablation} isolates refinement, routing, and execution under the same final-stage verification.

\begin{table}[!htb]
\centering
\caption{SpecWM ablations on the same 158-case WBench Navigation split. Scores are multiplied by 100; Nav. denotes NavScore. All variants use the same final-stage probe. Random routing matches route shares; synchronous execution disables draft-target overlap while retaining the full policy. \textbf{Bold}/\underline{underline} marks the best/second-best value per column.}
\label{tab:routing_ablation}
\scriptsize
\setlength{\tabcolsep}{1.5pt}
\renewcommand{\arraystretch}{1.12}
\begin{tabular*}{\columnwidth}{@{\extracolsep{\fill}}p{0.28\columnwidth}cccccc@{}}
\toprule
Variant & \shortstack{Speedup\\$\uparrow$} & \shortstack{HPSv3\\$\uparrow$} & \shortstack{Imag.\\$\uparrow$} & \shortstack{Aesth.\\$\uparrow$} & \shortstack{Spatial\\$\uparrow$} & \shortstack{Nav.\\$\uparrow$} \\
\midrule
\shortstack[l]{\textbf{SpecWM}\\\textbf{($3\to2\to1$)}} & 1.63$\times$ & \textbf{77.66} & \textbf{67.19} & \textbf{68.19} & \textbf{80.83} & \textbf{85.63} \\
\midrule
\multicolumn{7}{@{}l}{\textit{Refinement schedule}} \\
SpecWM (1-step) & \underline{1.84}$\times$ & 76.73 & 65.91 & 67.19 & 70.09 & 76.39 \\
SpecWM ($3\to1$) & 1.67$\times$ & 76.39 & 66.02 & 67.21 & \underline{76.90} & 81.49 \\
\midrule
\multicolumn{7}{@{}l}{\textit{Routing}} \\
w/o Refine (A/R) & 1.21$\times$ & 75.01 & 66.10 & 67.08 & 75.35 & 82.18 \\
w/o Reject (A/F) & \textbf{2.07}$\times$ & 75.89 & \underline{66.39} & 66.92 & 74.90 & 80.46 \\
Random routing & 0.95$\times$ & 73.71 & 64.91 & 65.28 & 67.64 & 75.98 \\
\midrule
\multicolumn{7}{@{}l}{\textit{Execution}} \\
w/o burst fallback & 1.45$\times$ & 76.49 & 65.81 & 66.87 & 75.62 & 81.39 \\
Synchronous execution & 1.29$\times$ & \underline{77.09} & 66.28 & \underline{67.50} & 76.59 & \underline{83.27} \\
\bottomrule
\end{tabular*}
\end{table}

\mainparagraph{Probe selection.}
\label{sec:exp_probe}
In a five-chunk pilot from one video, final-stage discrepancies span 0.1738--0.2698, a max/min ratio of 1.55 versus 1.22 and 1.30 at the first two stages. Separate optimized timings give 1.5\,s for one stage versus 4.4\,s for four-stage verification. The observed score variation, lower cost, and reusable candidate together motivate our choice of the final-stage probe.

\mainparagraph{Refinement schedule.}
The full schedule raises NavScore from fixed one-step refinement's 76.39 to 85.63. Removing the intermediate two-step correction lowers it to 81.49 and spatial consistency from 80.83 to 76.90, for only a small speedup increase (1.63$\times$ to 1.67$\times$). Thus, continued correction supports control and consistency after transitions.

\mainparagraph{Score-guided routing.}
Removing Refine lowers speedup to 1.21$\times$ and NavScore to 82.18; removing Reject raises speedup to 2.07$\times$ but lowers NavScore to 80.46. Thus, refinement reduces reliance on full regeneration, while retaining regeneration improves control. Moreover, random routing with matched shares gives 0.95$\times$ speedup and NavScore 75.98. Effective allocation depends on which chunks receive each action, not only on overall route shares.

\mainparagraph{Fallback and overlapped execution.}
\label{sec:exp_time}
Removing fallback or overlap lowers speedup from 1.63$\times$ to 1.45$\times$ or 1.29$\times$, respectively. Fallback skips repeated draft and probe costs during consecutive target generation, while overlap prepares the next draft during verification (Fig.~\ref{fig:overlap}). These gains persist with state restoration and re-proposal costs included, linking computation savings to lower generation latency.

\begin{table*}[!t]
\centering
\caption{LongSpecBench results on the 80-case evaluation split with 509-frame, multi-turn rollouts. Scores use $[0,100]$; generation-loop Speedup is relative to Target-only. WMQ and QES aggregate the displayed scores as defined in Sec.~\ref{sec:exp_longspec}. \textbf{Bold} and \underline{underline} mark the best and second-best values per column.}
\label{tab:longspec}
\scriptsize
\setlength{\tabcolsep}{3.0pt}
\renewcommand{\arraystretch}{1.15}
\begin{tabular*}{\textwidth}{@{\extracolsep{\fill}}lcccccccccc@{}}
\toprule
& Efficiency & \multicolumn{2}{c}{Visual $\uparrow$} & \multicolumn{2}{c}{Consistency $\uparrow$} & \multicolumn{3}{c}{Control $\uparrow$} & \multicolumn{2}{c}{Composite $\uparrow$} \\
\cmidrule(lr){2-2}\cmidrule(lr){3-4}\cmidrule(lr){5-6}\cmidrule(lr){7-9}\cmidrule(lr){10-11}
Method & Speedup $\uparrow$ & Imaging & Aesth. & Long-T. & Revisit & Dir. & Resp. & Cam. & WMQ & QES \\
\midrule
Target-only (14B) & 1.00$\times$ & \underline{67.62} & \textbf{64.47} & 90.47 & \textbf{76.77} & \underline{90.92} & \underline{83.41} & 18.87 & \underline{71.62} & 66.65 \\
Draft-only (1.3B) & \textbf{4.26}$\times$ & 66.95 & 64.10 & 88.83 & 66.40 & 85.42 & 72.00 & 16.60 & 67.13 & \underline{69.69} \\
Refine-all (3-step) & 1.08$\times$ & 67.46 & 63.71 & 91.05 & 74.75 & 90.79 & 82.83 & \underline{19.81} & 71.26 & 66.89 \\
\midrule
SDVG-style & 1.51$\times$ & 67.50 & 63.89 & 87.71 & 67.04 & 85.29 & 73.97 & 17.08 & 67.36 & 65.86 \\
SpecWM (3-step) & 1.21$\times$ & 66.78 & 64.03 & \underline{91.39} & 75.27 & 90.35 & 83.32 & 19.21 & 71.29 & 67.62 \\
\textbf{SpecWM ($3\to2\to1$)} & \underline{1.67}$\times$ & \textbf{68.26} & \underline{64.31} & \textbf{91.90} & \underline{76.56} & \textbf{91.89} & \textbf{83.61} & \textbf{22.03} & \textbf{72.41} & \textbf{70.32} \\
\bottomrule
\end{tabular*}
\end{table*}

\subsection{LongSpecBench: Long-Horizon Evaluation}
\label{sec:exp_longspec}

\mainparagraph{Benchmark design.}
We construct LongSpecBench to test sustained control and scene persistence over long rollouts with repeated turns and revisits. Its 100 initial scenes come from WBench~\cite{wbench} (30), WorldMark~\cite{worldmark} (30), and TrajectoryBench~\cite{logo} (40). The first 20 cases set $\tau=0.2395$ and $\tau_r=0.2716$ using the same 70th/90th-percentile rule; the remaining 80 form the evaluation split. Each rollout contains ten to twelve action turns and revisits over 509 frames at 16\,fps (about 32 seconds, 32 chunks), with the same input-dependent resolution as WBench.

\mainparagraph{Evaluation dimensions.}
Seven metrics measure visual quality (Imaging, Aesth.), temporal and revisit consistency (Long-T., Revisit), and control through motion direction (Dir.), response to changes (Resp.), and camera following (Cam.). Cam. scores trajectory following on $[0,100]$ using poses estimated by MegaSaM~\cite{megasam}.

\mainparagraph{Composite evaluation metrics.}
World-model quality is $\mathrm{WMQ}=0.30Q+0.35C+0.35U$, where $Q$, $C$, and $U$ average the visual-quality, consistency, and control groups. For Speedup $S$, efficiency is $E=100S/(1+S)$, giving the quality-efficiency score $\mathrm{QES}=100(\mathrm{WMQ}/100)^{0.80}(E/100)^{0.20}$. Both composites use the method-level scores in Table~\ref{tab:longspec}, with equal weights within groups and rounding to two decimals only after calculation.

\mainparagraph{Long-horizon results.}
SpecWM preserves near-target revisit consistency (76.56 versus 76.77) at 1.67$\times$ speedup, while leading all three control metrics (Table~\ref{tab:longspec}). It closes most of the draft's revisit gap, from 66.40 to 76.56. These results extend the shorter-rollout findings: selective correction sustains control and scene persistence through repeated turns and returns. Its leading WMQ (72.41) and QES (70.32) summarize this quality-efficiency trade-off.

\noindent\begin{minipage}{\columnwidth}
\centering
\captionsetup{hypcap=false}
\includegraphics[width=\columnwidth]{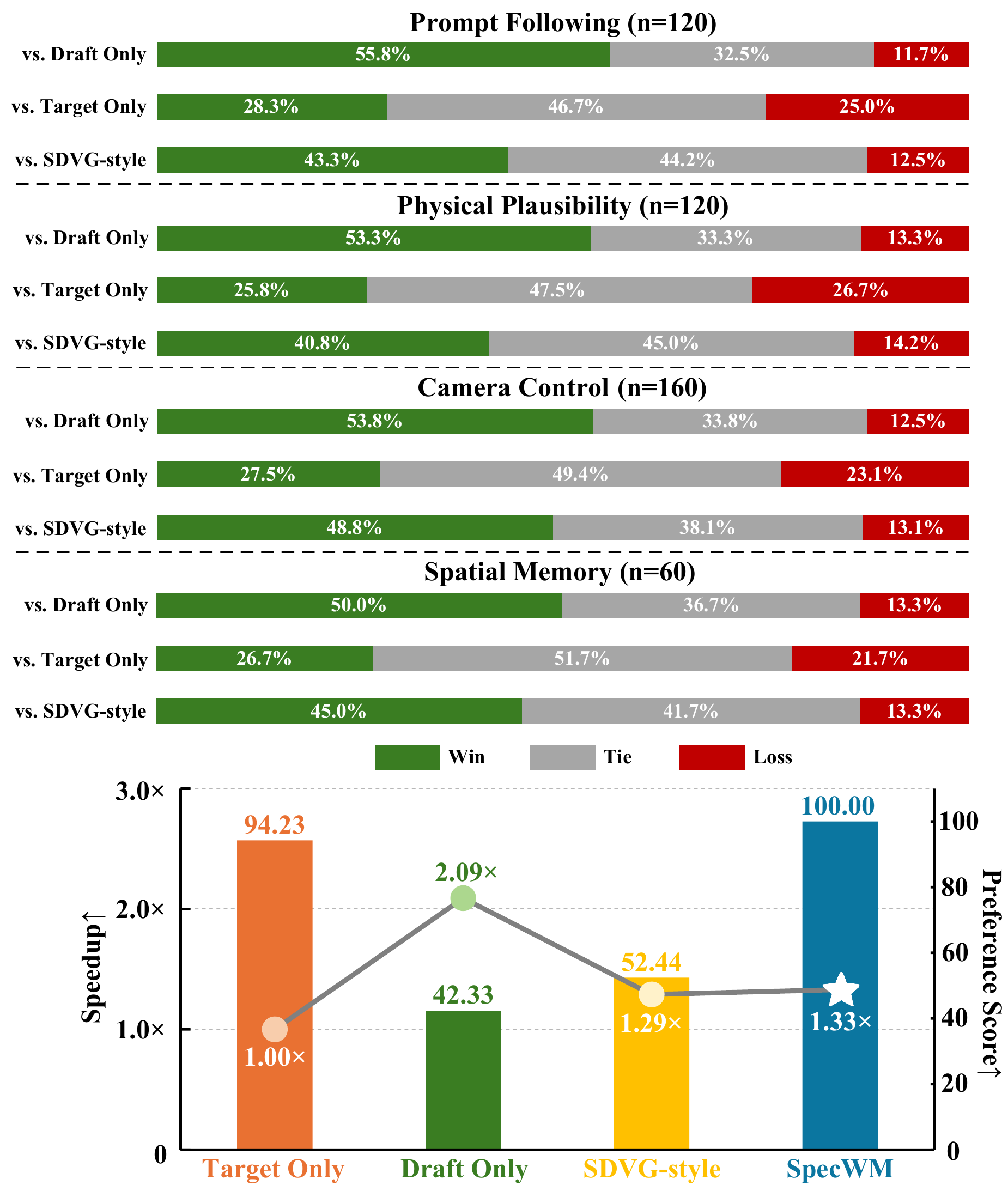}
\par\vspace{3.18pt}
\captionof{figure}{\textbf{Hy-Nomas Arena evaluation.} Win/Tie/Loss percentages are from SpecWM's perspective; $n$ denotes cases per dimension. The four-step draft and eight-step target share 8B weights. Bottom: relative preference scores (bars) and generation-loop speedup over Target-only (line). Scores equally weight the four dimensions, count ties as half-wins, and normalize derived preference strengths to a maximum of 100. The supplementary material gives the calculation from the original pairwise counts.}
\label{fig:hynomas_arena}
\end{minipage}
\par

\subsection{Pairing by Denoising Budget}
\label{sec:exp_hynomas}

\mainparagraph{Shared-weight pairing.}
We evaluate Hy-Nomas~\cite{hynomas} with the same distilled 8B weights, using four sampling steps for the draft and eight for the target. A fixed terminal probe provides a score and correction, while transition-aware refinement selects correction intervals on the native grid. Pairwise human evaluation uses the official 280-case Hy-Nomas-Bench; other settings follow native defaults. Sampling and evaluation details appear in the supplementary material.

\mainparagraph{Human preference and efficiency.}
SpecWM achieves near-target human preference at 1.33$\times$ generation-loop speedup (Fig.~\ref{fig:hynomas_arena}). Against Target-only, ties dominate each dimension, with comparable win and loss rates. SpecWM also records positive win-loss margins against Draft-only and SDVG-style across all four dimensions. Compared with SDVG-style, higher preference in camera control and spatial memory accompanies faster generation. Together, relative preference scores and speedups summarize SpecWM's balance between human preference and efficiency.

\mainparagraph{Across backbones and sampling budgets.}
Together with LingBot, these results validate SpecWM across backbones and pairing strategies. Specifically, LingBot uses different model scales, whereas Hy-Nomas varies the sampling budget within one network. Finer denoising steps provide corrections without increasing model capacity. SpecWM accelerates generation even with identical model weights. Native sampling budgets offer another way to form draft-target pairs without training a separate draft model.

\Needspace{3\baselineskip}
\section{Conclusion}
\label{sec:conclusion}
We present SpecWM, a unified, training-free inference framework that turns verification into refinement for streaming world models. The target's latent velocity prediction supplies both a verification signal and a reusable correction, reducing reliance on costly regeneration. Our experiments reveal complementary benefits of short and long refinements for perceptual quality and trajectory control. Transition-aware refinement combines these benefits, achieving the best overall quality-speed-control trade-off among the evaluated methods on WBench Navigation and our LongSpecBench. Hy-Nomas further demonstrates effective pairing through different denoising budgets with shared weights. Together, these findings support verification-guided refinement across model-scale and sampling-budget differences. Future work will extend draft-target pairing through cross-model latent alignment.

\clearpage
\section*{LLM Usage Statement}
We used OpenAI GPT to refine our writing, related-work discussions, experimental analyses, figures, and data preparation. We also used it to generate research code for implementation and experiments. We take full responsibility for the accuracy and integrity of all submitted content.

{\small
\bibliographystyle{ieeenat_fullname}
\bibliography{main}
}
\end{document}